\documentclass[10pt,conference]{IEEEtran}
\usepackage{cite}
\usepackage{amsmath,amssymb,amsfonts}
\usepackage{algorithmic}
\usepackage{graphicx}
\usepackage{textcomp}
\usepackage{xcolor}
\usepackage{subcaption}
\usepackage{hyperref}

\begin{document}
\title{Road Damage Detection using Deep Ensemble Learning}

\author{\IEEEauthorblockN{Keval Doshi}
\IEEEauthorblockA{\textit{Department of Electrical Engineering} \\
\textit{University of South Florida}\\
Tampa, USA \\
kevaldoshi@usf.edu
}
\and
\IEEEauthorblockN{Yasin Yilmaz* \thanks{corresponding author}}
\IEEEauthorblockA{\textit{Department of Electrical Engineering} \\
\textit{University of South Florida}\\
Tampa, USA \\
yasiny@usf.edu
}}

\maketitle

\begin{abstract}
Road damage detection is critical for the maintenance of a road, which traditionally has been performed using expensive high-performance sensors. With the recent advances in technology, especially in computer vision, it is now possible to detect and categorize different types of road damages, which can facilitate efficient maintenance and resource management. In this work, we present an ensemble model for efficient detection and classification of road damages, which we have submitted to the IEEE BigData Cup Challenge 2020. Our solution utilizes a state-of-the-art object detector known as You Only Look Once (YOLO-v4), which is trained on images of various types of road damages from Czech, Japan and India. Our ensemble approach was extensively tested with several different model versions and it was able to achieve an F1 score of 0.628 on the test 1 dataset and 0.6358 on the test 2 dataset.  
\end{abstract}


\section{Introduction}
Road infrastructure is a crucial public asset as it contributes to economic development and growth while bringing critical social benefits \cite{arya2020transfer}. Specifically, road maintenance is pivotal in the socioeconomic development and for a smooth continuation of day-to-day operations in a country. However, it is a challenge for governments and state agencies to constantly perform pavement condition surveys. While several states in the U.S. employ some sort of semi-automated methods such as using road survey vehicles equipped with a multitude of sensors to evaluate pavement conditions and deterioration, using dedicated vehicles and imaging equipment is often expensive and unaffordable to the local agencies. Moreover, in several developing countries, the process is completely manual and involves long hours of visually inspecting the condition of roads by conducting a windshield survey from a slow moving vehicle on a regular basis. Also, evaluating the state of the structural damage is subjective and requires experts to judge the extent of damage. Due to the negative trend in infrastructure maintenance and management, it is clear that more efficient and sophisticated infrastructure maintenance methods are urgently required. Therefore, several research and commercial efforts have been conducted to aid government agencies to automate the road inspection and sample collection process, making use of technologies with varied degrees of complexity \cite{medina2014enhanced}. 

With the advances in deep learning and particularly image processing, several low-cost methods that involve collecting images and leveraging deep learning-based algorithms have been proposed \cite{Maeda_2018,alfarrarjeh2018deep,wang2018road}. Several earlier works only focused on detecting the existence of road damage rather than recognizing its type. However, it is difficult to compare the models proposed in earlier works as they use different datasets which are significantly different from each other. Since it is also critical to differentiate between different damage types, in \cite{Maeda_2018} Maeda et al. proposed a comprehensive dataset consisting of 9053 images and 8 damage types. More recently, they extended the dataset by including images from Czech, India and Japan. By leveraging the proposed dataset and the benchmark algorithms, it is now possible to propose new modifications and improvements. In this work, we propose to use a one-stage detector called "You Only Look Once" (YOLO-v4) as it is capable of achieving high accuracy at a reasonable computational complexity.    

The remainder of this paper is organized as follows. In Section II, we discuss the related works. In Section III, the proposed approach and the results are presented. In Section IV, we discuss the future scope and suggest improvements. Concluding remarks are given in Section V.

\section{Related Work}
Road damage detection and classification has been an active area of research for the computer vision and civil engineering communities. Although there had been a lot of work, e.g., \cite{zalama2014road,nishikawa2012concrete}, on applying image processing approaches for the problem, \cite{zhang2016road} was the first to apply CNNs to road damage detection. Since then, other works \cite{pereira2018deep,zhang2017automated,akarsu2016fast} have focused on using deep learning for crack detection in road images. However, other than \cite{Maeda_2018}, most of these works have been limited to detection of one particular type of damage or classifying images based on damage type. In \cite{Maeda_2018}, a new dataset was proposed named "RDD 2018", which consisted of 8 different damage categories. As a result, the underlying data, method, and models have gained wide attention from researchers all over the world \cite{arya2020transfer}. Specifically, a technical challenge was organized in December 2018 as a part of the IEEE Big Data Conference held at Seattle, USA, which utilized this data for evaluating the performance of several models for road condition monitoring. In total, 59 teams participated in the challenge from 14 different countries\cite{manikandan2018varying,alfarrarjeh2018deep}. Another work by Du et al. \cite{du2020pavement}  uses a dataset of 45,788 road images collected from Shanghai and utilize YOLO model for detecting and classifying pavement distresses. However, in that dataset, the images were collected using an industrial high-resolution camera, unlike our work, which is based on the less expensive smartphone-based images. In \cite{majidifard2020pavement}, Majidifard et al. used Google street view images considering both top-down as well as wide-view options for classification and densification of pavement distresses collected from 22 different pavement sections in the United States. However, the size of the dataset used is limited to only 7237 images. Ideally, more than 5000 labeled images are generally required for each class for an image processing-based classification task to provide satisfactory results.
\section{Methodology and Results}\label{eval}

\subsection{Dataset and Evaluation Metric}
\label{sec:dataset}

The dataset used in this work was proposed in \cite{arya2020transfer} and consists of 26620 labeled road damage images belonging to 4 classes acquired from a smartphone camera from India, Czech and Japan. Specifically, 3595 images were collected from Czech, 9892 from India and 13133 from Japan, consisting of a total of 31343 bounding boxes. The damage types included D00 (longitudinal linear crack), D10 (lateral linear crack), D20 (alligator crack) and D40 (pothole damage). The models were evaluated on two datasets consisted of images randomly picked from each country. A prediction was considered as correct if
\begin{itemize}
\item the predicted bounding box had the same class label as the ground truth bounding box, and
\item the predicted bounding box had over 50\% Intersection over Union (IoU) in with the ground truth bounding box. 
\end{itemize}
The final metric used for evaluation is the $F_1$ score. The $F_1$ score measures accuracy using the statistics of precision $p$ and recall $r$. Precision is the ratio of true positives $(tp)$ to the total number of bounding boxes detected $(tp + fp)$ while recall is the ratio of true positives to the actual number of bounding boxes  $(tp + fn)$. The $F_1$ score is given by:

\begin{equation}
    F_1 = 2 * \frac{p*r}{p+r}, \text{where  } p = \frac{tp}{tp+fp} \text{and  } r = \frac{tp}{tp+fn}
\end{equation}

\vspace{1pt}

\subsection{Deep Ensemble Learning}
\label{DEL}

An object detection algorithm deals with detecting semantic objects and visual content belonging to a certain class from a digital image. With the advances in deep neural networks, several Convolutional Neural Network (CNN) based object detection algorithms have been proposed. The first one was the Region of CNN features (R-CNN) method \cite{girshick2014rich}, which proposed to perform object detection via two steps: object region proposal and classification. The first step generates multiple regions by using a selective search, which are then input to a CNN classifier. However, due to its inherent computational complexity, several optimized versions of R-CNN were proposed such as the Fast R-CNN \cite{ren2015faster} algorithm. More recently, an algorithm known as "You Only Look Once” (YOLO) \cite{redmon2016you} was proposed, which combined the two steps from R-CNN algorithm and significantly reduced the computational complexity. YOLO uses a CNN which inherently decides regions from the image and outputs probabilities for each of them. Hence, it is able to achieve a significant speedup as compared to R-CNN based algorithms and can be used for real-time processing as well. The goal of this work is to improve upon the real-time detection capabilities for road damage detection, hence we use YOLO as our base model. Ensemble methods, which combine the predictions from various models, have been successfully employed in various machine learning tasks to improve the accuracy. In this work, we use an ensemble of YOLO-v4 models trained for different number of iterations and different resolutions. More details about the model selection and implementation can be found here\footnote{\url{https://github.com/kevaldoshi17/IEEE-Big-Data-2020}}. We present the model performance in Table \ref{tab2}. 

\subsection{Results} 

\begin{figure*}[thb]
\minipage{0.32\textwidth}
  \includegraphics[width=\linewidth]{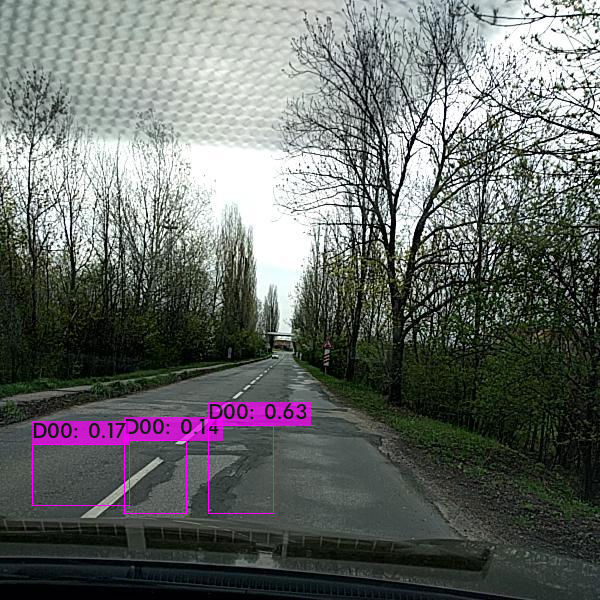}
\label{f:Testbed}
\endminipage\hfill
\minipage{0.32\textwidth}
  \includegraphics[width=\linewidth]{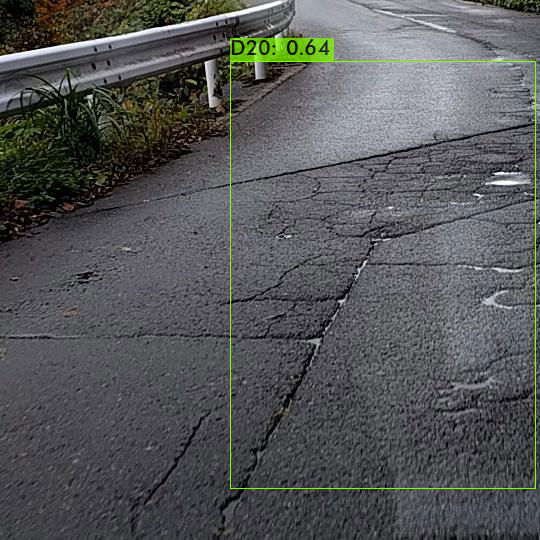}
  \label{f:http_fig}
\endminipage\hfill
\minipage{0.32\textwidth}%
  \includegraphics[width=\linewidth]{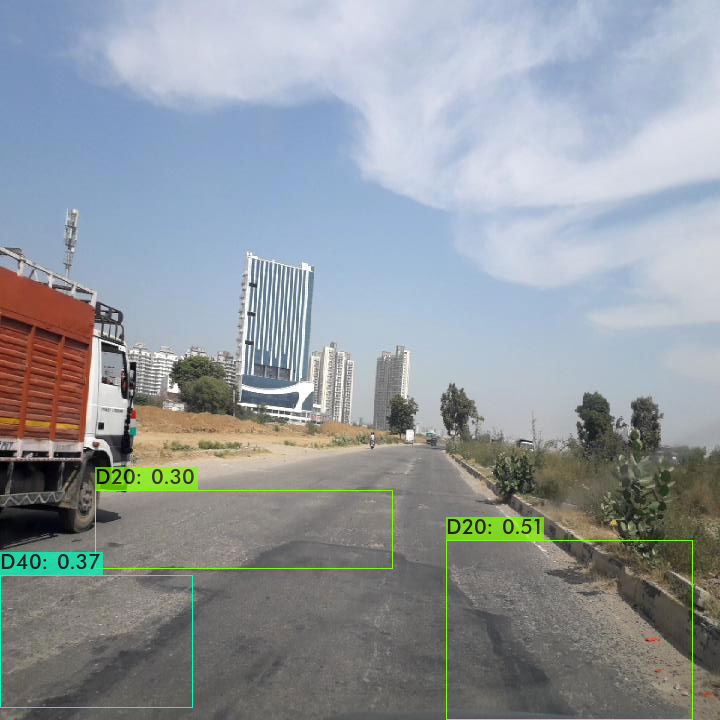}
  \label{f:miti}
\endminipage
\caption{Detection Results from a single YOLO-v4 model for different areas under varying lighting conditions and viewpoints in Czech, Japan, and India respectively.} 
\label{fig:animals}
\end{figure*}

\begin{figure*}
\centerline{\includegraphics[width=1\textwidth]{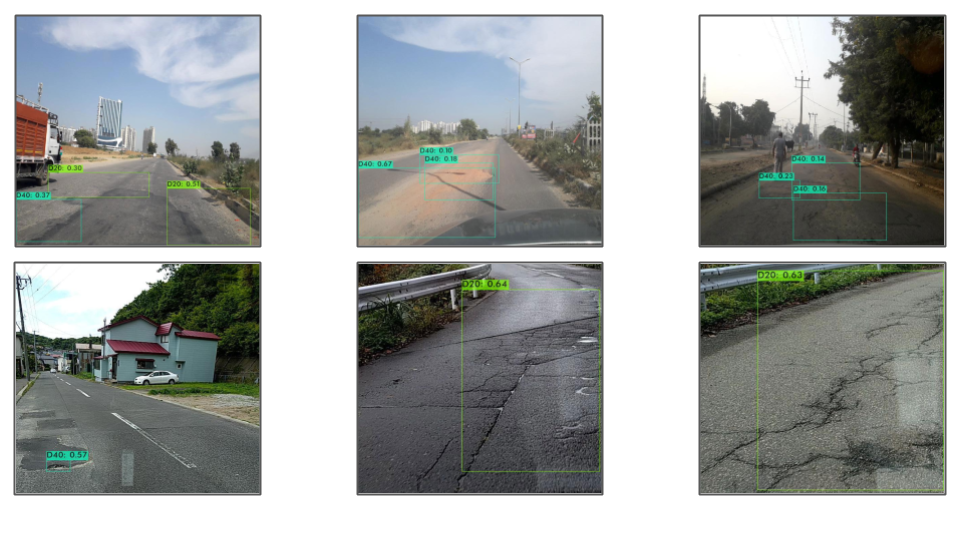}}
\caption{Labels predicted for India (top row) and Japan (bottom row) using the model trained on data from the two countries.  \label{fig11a}}
\end{figure*}

\begin{figure*}
\centerline{\includegraphics[width=1\textwidth]{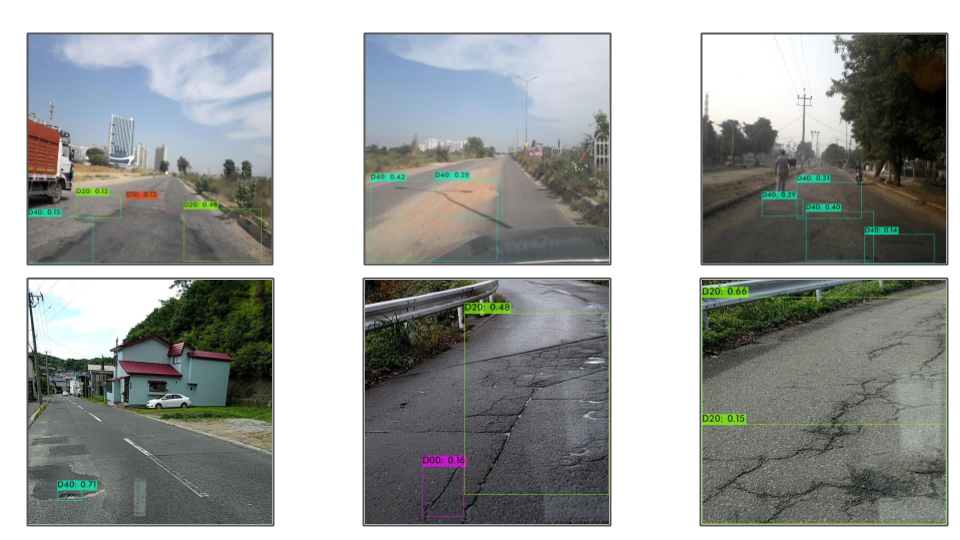}}
\caption{Labels predicted for India (top row) and Japan (bottom row) using the model trained on all the three countries. \label{fig11b}}
\end{figure*}

Fig. \ref{fig:animals} shows the detection results from a single YOLO-v4 model under varying conditions. 
In Fig. \ref{fig11a}, we show some detection results for YOLO models trained on data from Japan and India. Whereas, in Fig. \ref{fig11b}, training data from Japan, India, and Czech are used. The models trained on data from all the countries seem to perform better than the models trained on data from Japan and India only, which is in contrast to the results presented in \cite{arya2020transfer}. 

Furthermore, selection of the input image size considerably affects the detection performance. Since YOLO requires the input image resolution to be a multiple of 32, we focused on two specific sizes, 416 and 608. However, as opposed to common perception, increasing the resolution of the image decreased the performance of the base model, 
as shown in Table \ref{tab2}.

We evaluated the performance of the proposed models by using the platform provided by the organizers of IEEE BigData Cup Challenge 2020. As described in Section \ref{sec:dataset}, the bounding boxes whose class label matched with the ground truth were selected and then those with a greater than 50\% IoU were picked. Finally the $F_1$ Score for these boxes was calculated. 


\begin{table}[htbp]
\caption{Model Comparison}
\begin{center}
\begin{tabular}{|c|c|c|}
\hline
\textbf{Model}&\multicolumn{2}{|c|}{\textbf{$F_1$} \textbf{Score} $^{\mathrm{a}}$} \\
\cline{2-3} 
\textbf{Name} & \textbf{\textit{Dataset: Test 1}}& \textbf{\textit{Dataset: Test 2}} \\
\hline
Yolo-v4 (416x416) & 0.5193 & 0.5137 \\
\hline
Yolo-v4 (608x608) & 0.5122 & 0.5012 \\
\hline
Ensemble (5 models) & 0.5321 & 0.5226 \\
\hline
Ensemble (15 models) & 0.6091 & 0.5983 \\
\hline
Ensemble (25 models) & 0.6102 & 0.6297 \\
\hline
Ensemble (30 models) & 0.6275 & 0.6358 \\
\hline
\multicolumn{3}{l}{$^{\mathrm{a}}$ Only correctly classified boxes were considered from all predictions}
\end{tabular}
\label{tab2}
\end{center}
\end{table}

Table \ref{tab3} presents our final scores among other teams. Using the ensemble of 30 models shown in Table \ref{tab2}, we achieved the second position overall.

\begin{table}[htbp]
\caption{Result Comparison}
\begin{center}
\begin{tabular}{|c|c|c|}
\hline
\textbf{Teaml}&\multicolumn{2}{|c|}{\textbf{F1 Score}} \\
\cline{2-3} 
\textbf{Name} & \textbf{\textit{Dataset: Test 1}}& \textbf{\textit{Dataset: Test 2}} \\
\hline
IMSC & 0.674 & 0.666 \\
\hline
SIS Lab (Ours) & 0.628 & 0.6358 \\
\hline
DD Vison & 0.665 & 0.6219 \\
\hline
IME\&GIICS & 0.599 & 0.6018 \\
\hline
mynotwo & 0.588 & 0.5963 \\
\hline
titan mu & 0.581 & 0.575 \\
\hline
\end{tabular}
\label{tab3}
\end{center}
\end{table}

\section{Future Scope}

While this research serves as a baseline for road damage detection, there are several improvements that can be made. First, extracting and combining informative features from smartphones would allow a detector to make better decisions. For example, data fusion using accelerometer readings and audio would help significantly reduce the number of false alarms. Secondly, including video as an input to the detector would allow for sequential detection algorithms, which can lead to fewer false alarms. 
An additional increase in coverage 
could be achieved by installing the road damage detection system on smartphones and cameras mounted on the vehicles operated by municipalities, such as public transport or waste collection vehicles. 

With new test streams for video, audio, and accelerometer data, we can use obtain joint features by applying the same object detectors trained in this research followed by a forward propagation over a set of trained deep generative models (e.g., variational autoencoders). Note that the proposed framework is an end-to-end joint feature extraction methodology. The inputs are video, audio, accelerometer data streams, and the outputs are features of joint  representations. These features can then be used in a sequential anomaly detection and classification framework. 

Furthermore, continual learning is an important extension that can be further explored to improve the detection performance and learn new types of road damages without retraining the entire model from scratch. 

Finally, adapting to new road environments is a challenging task. By using meta learning algorithms, it could be possible to propose a single standardized model that is applicable globally or at least to a set of countries having similar road conditions.

\section{Conclusion}

In this paper, we proposed an ensemble model for the road damage detection task. By utilizing the state-of-the-art YOLOv-4 object detector as the base model, we were able to achieve a competitive performance. 
Specifically on the road damage datasets provided by the IEEE BigData Cup Challenge 2020, we got the second rank in the competition. We further discussed several possible extensions for improving the dataset and the road damage detection field in general.  

\bibliography{references}{}
\bibliographystyle{ieeetr}

\end{document}